\documentclass[conference]{IEEEtran}
\pdfoutput=1

\IEEEoverridecommandlockouts
\usepackage{cite}
\usepackage{amsmath,amssymb,amsfonts}
\usepackage{algorithmic}
\usepackage{graphicx}
\usepackage[caption=false, font=footnotesize]{subfig}
\usepackage{textcomp}
\usepackage{xcolor}
\def\BibTeX{{\rm B\kern-.05em{\sc i\kern-.025em b}\kern-.08em
    T\kern-.1667em\lower.7ex\hbox{E}\kern-.125emX}}

\usepackage[svgpath=./fig/]{svg}
\usepackage{multirow}
\usepackage{float}

\usepackage{tikz}
\usetikzlibrary{shapes,arrows,positioning,calc}
\tikzset{
    arrow/.style={->}
}

\usepackage{amsmath,bm,times}

\newcommand\copyrighttext{%
  \footnotesize \textsuperscript{\textcopyright} 2020 IEEE. Personal use of this material is permitted.  Permission from IEEE must be obtained for all other uses, in any current or future media, including reprinting/republishing this material for advertising or promotional purposes, creating new collective works, for resale or redistribution to servers or lists, or reuse of any copyrighted component of this work in other works.}
\newcommand\copyrightnotice{%
\begin{tikzpicture}[remember picture,overlay]
\node[anchor=south,yshift=0pt] at (current page.south) {\fbox{\parbox{\dimexpr\textwidth-\fboxsep-\fboxrule\relax}{\copyrighttext}}};
\end{tikzpicture}%
}

\begin{document}

\title{Evaluation of 3D CNN Semantic Mapping for Rover Navigation\\
\thanks{Project BIRD181070 funded by the program BIRD 2018 sponsored by the University of Padova.}
}

\author{\IEEEauthorblockN{Sebastiano Chiodini$^{(1)(*)}$, Luca Torresin$^{(1)}$, Marco Pertile$^{(1)}$ and Stefano Debei$^{(1)}$}
	\IEEEauthorblockA{
		$^{(1)}$ Department of Industrial Engineering, University of Padova, Via Venezia 1, 35131 Padova, Italy\\
		$^{(*)}$ \textit{corresponding author:} sebastiano.chiodini@unipd.it
}}

\maketitle

\copyrightnotice{}

\begin{abstract}
Terrain assessment is a key aspect for autonomous exploration rovers, surrounding environment recognition is required for multiple purposes, such as optimal trajectory planning and autonomous target identification. In this work we present a technique to generate accurate three-dimensional semantic maps for Martian environment. The algorithm uses as input a stereo image acquired by a camera mounted on a rover. Firstly, images are labeled with DeepLabv3+, which is an encoder-decoder Convolutional  Neural Networl (CNN). Then, the labels obtained by the semantic segmentation are combined to stereo depth-maps in a Voxel representation. We evaluate our approach on the ESA Katwijk Beach Planetary Rover Dataset.
\end{abstract}

\begin{IEEEkeywords}
convolutional neural network, semantic segmentation, Martian environment
\end{IEEEkeywords}

\section{Introduction}

Autonomous navigation of mobile robots can enormously increase the scientific output of an in-situ exploration mission, as an example the distance traveled for each sol by the NASA MSL rover has increased from few meters up to 100 m \cite{bajracharya2008autonomy,heverly2013traverse}. Navigation problem is generally divided into the following sub-tasks: localization, terrain assessment and path planning. Navigating on Mars has peculiarities that we do not find in other autonomous robotics applications like car driving in urban environment. Firstly, localization suffers of lack of GPS signal and it needs vision techniques such as Visual Odometry (see \cite{pertile2016uncertainty}, \cite{giubilato2019}) to estimate rover ego-motion. Secondly, terrain assessment deals with an unstructured environment that is characterized by pointed rocks (large or small), sand and bedrocks \cite{rothrock2016spoc}. 

\begin{figure}[t]
\centering
\includegraphics[width=1\linewidth]{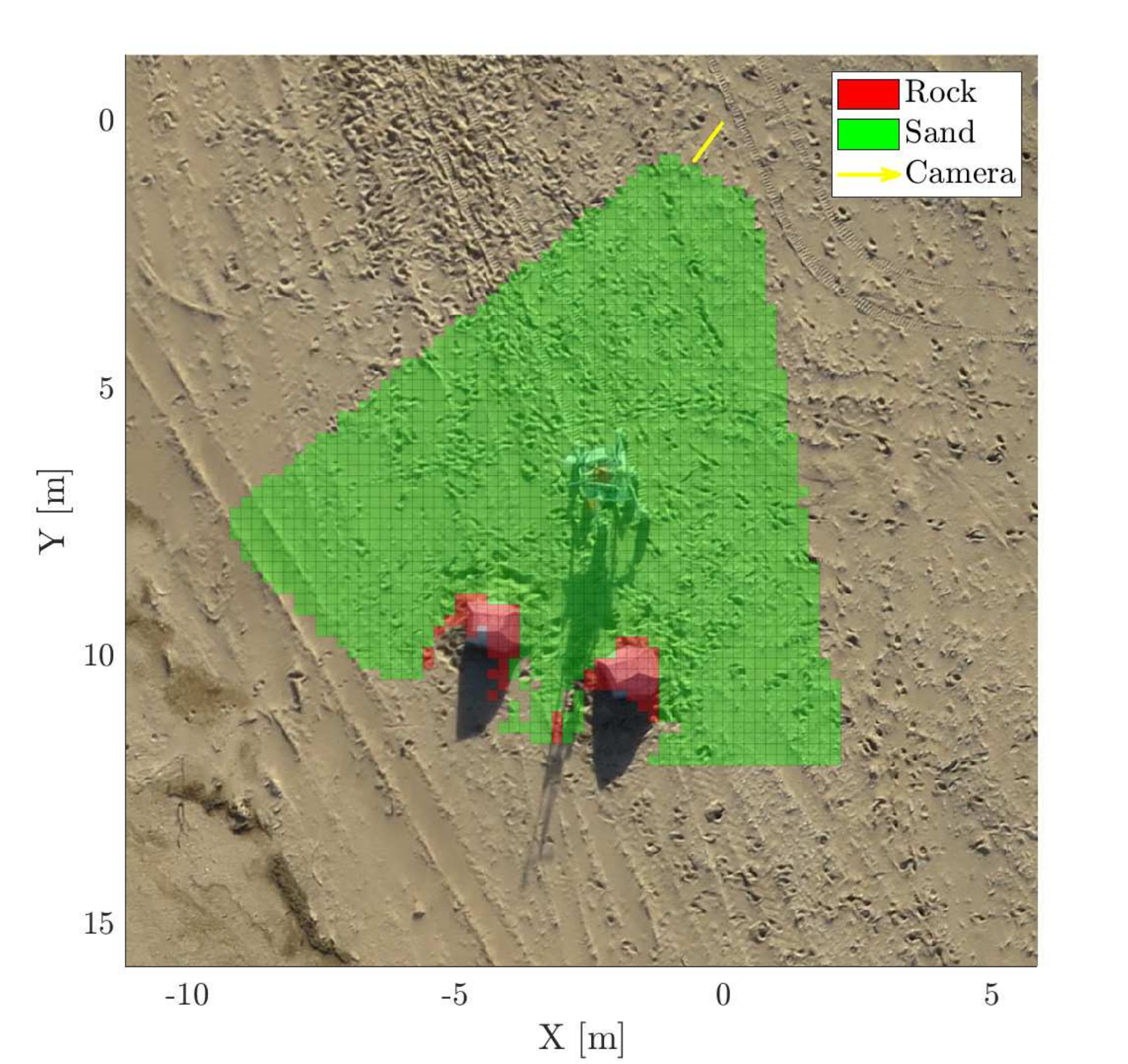}

\caption{Terrain assessment based on semantic segmentation of a stereo-camera point cloud. The classifier, which is based on encoder-decoder network DeepLabv3+, segment the scene between rocks and sand.}
\label{fig::image_segmentation_map}
\end{figure}

One of the first autonomous terrain assessment systems for Martian environment was implemented on-board the NASA MER rovers: the GESTALT (Grid-based Estimation of Surface Traversability Applied to Local Terrain) system \cite{maimone2007overview}. GESTALT was able to detect geometric hazards such as rock, ditches and cliffs by processing the 3D points generated by the rover stereo-images, it looked mainly at geometric characteristics such as steps, slopes and terrain roughness. However, geometric point cloud processing could be computationally expensive. For this reason the possibility of directly merging the depth map with labels, called semantic mapping, reduces the terrain assessment computational time, \figurename{\ref{fig::image_segmentation_map}} shows an example of a terrain assessment technique based on semantic mapping. Moreover, image labelling adds information that, otherwise, cannot be deduced from the three-dimensional model alone, such as sandy or slippery terrain detection \cite{angelova2007slip}. 

\pgfdeclarelayer{background}
\pgfdeclarelayer{foreground}
\pgfsetlayers{background,main,foreground}

\tikzstyle{sensor}=[draw, fill=blue!20, text width=7.5em, 
    text centered, minimum height=7em, rounded corners]
\tikzstyle{frontend}=[draw, fill=red!20, text width=7.5em, 
    text centered, minimum height=2.5em, rounded corners]
\tikzstyle{labeler}=[draw, fill=gray!20, text width=7.5em, 
    text centered, minimum height=2.5em, rounded corners]
\tikzstyle{map}=[draw, fill=green!20, text width=7.5em, 
    text centered, minimum height=2.5em,  rounded corners]
\tikzstyle{ann} = [above, text width=5em]
\tikzstyle{naveqs} = [sensor, text width=6em, fill=red!20, 
    minimum height=12em, rounded corners]
\tikzstyle{line} = [draw, -latex']
\def\blockdist{2.3}
\def\edgedist{2.5}

\begin{figure*}[h]
    \centering

    \includegraphics[width=1\linewidth]{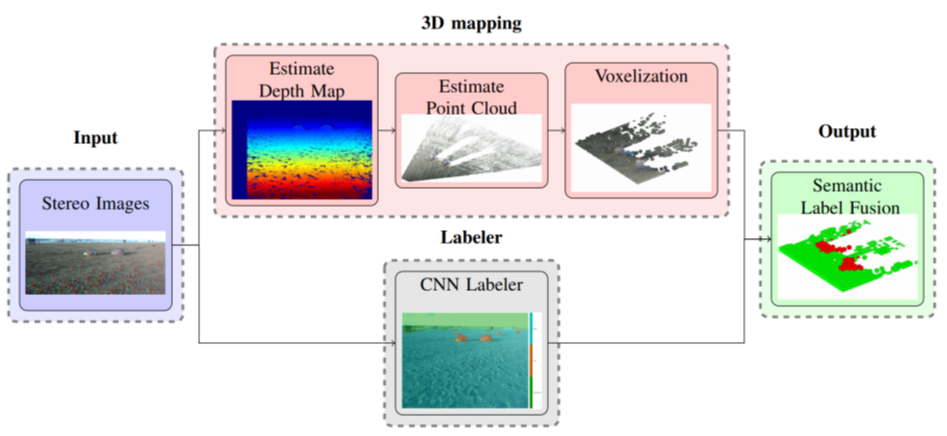}

    \caption{Scheme overview of the proposed terrain assessment algorithm. The algorithm take as input a rectified stereo image, which is used to estimate the depth map and the point cloud of the scene. Afterwards, the point cloud are changed to a voxel grid in order to obtain a volumetric representation of the terrain. Simultaneously a label is associated to each pixel of the stereo image with a CNN labeler (DeepLabv3+). Finally, the outputs of the labbeler are merged to the voxel grid to obtain a 3D semantic model of the scene.}
    \label{fig:pipeline}
\end{figure*}

In literature it is possible to find many works about semantic mapping in urban environment, as an example \cite{sengupta2013urban} generates dense 3D reconstruction with associated semantic labellings from stereo-camera images and \cite{jeong2018multimodal} proposes a multimodal sensor-based semantic 3D mapping system using a 3D Lidar combined with a camera. However, these techniques have rarely been applied to the case of planetary exploration and environments with a low number of features. In \cite{chiodini2020retrieving} is presented a method for fusing range measurements with monocular Visual Odometry mapping.

In this paper is proposed a terrain assessment method for Martian environment based on semantic mapping which uses as input a set of stereo-images. The three-dimensional map is represented with a voxel grid, in order to improve the computational efficiency, and the labelling is performed with the state-of-the-art Convolutional Neural Network labeler DeepLabv3+ \cite{chen2018encoder}. In order to evaluate the metrological performances of the proposed method images from the ESA Katwijk Beach Planetary Rover Dataset \cite{hewitt2018katwijk} have been processed. The dataset was created for the validation of localization and navigation algorithms in Martian-like environment and it provides trajectory and map ground truth. The dataset sequences are challenging by the presence of artifacts produced by sunlight.

The major contributions of this work are:
\begin{itemize}
    \item The testing of semantic mapping techniques into a Martian analogous environment.
    \item The application of the relatively new convolutional neural network architecture DeepLabv3+ for terrain assessment.  
    \item The comparison of the proposed method with a method based on geometric  point  cloud  processing \cite{torr2000mlesac}.
    \item The performances evaluation using a manually segmented ortomosaic map as reference. 
\end{itemize}
The paper is divided in the following sections: Section II describes the proposed terrain assessment method, Section III shows the algorithm performances in a Martian analogous environment and in Section IV the conclusions of this paper are reported.

\section{Method}

The algorithm takes as input a rectified stereo image $(I_l,I_r)$, which is used to estimate the depth map and the point cloud of the scene. Afterwards, the point cloud is transformed into a voxel grid in order to obtain a volumetric representation of the terrain. Simultaneously a label is associated to each pixel of the stereo image with a CNN labeler (DeepLabv3+). Finally, the outputs of the labbeler are merged to the voxel grid by following  the  standard  Bayes’  update rule to obtain a 3D semantic model of the scene. \figurename{\ref{fig:pipeline}} shows the proposed terrain assessment method pipeline.

\subsection{3D mapping}

First of all the stereo-images need to be stereo-rectified. Then, once the stereo-images have been stereo-rectified it is possible to compute the disparity map of the scene. We use the Semi-Global Block Matching algorithm \cite{hirschmuller2005accurate}, this function computes the scene disparity by comparing the sum of absolute differences (SAD) for each block of pixels in the image and forces similar disparity on neighboring blocks. By knowing the disparity map and the calibration parameters is possible to estimate the depth of the scene and compute the related point cloud ($\mathbf{X}_{j} = [X_{j}, Y_{j}, Z_{j}]$), where $j=1,\dots,n$ with $n$ number of pixels with an associated disparity. The 3D points coordinates in the left camera frame are given by: 
\begin{equation}
    \begin{split}
X_{j} & = \frac{u_l Z_{j}}{f} \\
Y_{j} & = \frac{v_l Z_{j}}{f} \\
Z_{j} & = \frac{b f}{u_l - u_r}
\end{split}
\end{equation}
where $(u_l,\, v_l)$ and $(u_r,\, v_r)$ are respectively the pixel coordinates in the left and right image, $b$ is the camera baseline and $f$ is the camera focal length.
At the end the point cloud is converted in a three dimensional voxel grid   follow  the  standard  Bayes’ update rule. In the experimental part a grid resolution of $0.2\times0.2\times0.2$ m has been considered. 

\newpage

\subsection{Labeling}

The purpose of image labeling part is to attribute a label to each pixel of the image
$(u_l,\, v_l)$ that are used to build the map. Image labeling is performed by using a Convolutional Neural Network (CNN) technique, as CNNs have demonstrated significant improvement in semantic segmentation tasks. Between the state-of-the art methods (FCN \cite{long2015fully}, SegNet \cite{badrinarayanan2017segnet} and U-net \cite{ronneberger2015u}) we choose DeepLabv3+ \cite{chen2018encoder} because it showed better performances in preliminary tests. DeepLabv3+ which has an encoder-decoder structure is a pre-trained network. Pre-training has been performed on ImageNet, which is a dataset containing more than one million images divided into 1000 classes. This allows to train the deeper layer of the network with the most characteristic feature. Instead the final part of the training phase is application dependant, and is used to train the outer layer of the network.

Due to the type of the dataset environment only three labels have been selected: \textit{sand}, \textit{rocks}, \textit{background}. \figurename{\ref{fig::image_segmentation}} shows an example of image semantic segmentation application on the testing dataset. Taking advantage of a pre-trained network, we limited the number of images. In order to improve the training accuracy ad robustness to image variation the following techniques of data augmentation have been applied: cropping, mirroring and resizing. The number of images used for training is 400: 50 original images and 350 obtained with data augmentation.

\subsection{Data fusion}

Each 3D point of the point cloud ($\mathbf{X}_{j}$) is associate with a pixel of the image $(u_l,\, v_l)$. For the label fusion to the occupancy grid map we follow the standard Bayes' update rule \cite{thrun2002probabilistic}. Assuming that the grid map cells are independent the one to each other, the probability belief of a single cell $p(m_{x,y}|z_{1:j})$ given a series of rock observations $z_{1:j}$ is:
\begin{equation}
    p(m_{x,y}|z_{1:j})=\frac{p(m_{x,y}|z_{j})p(z_{j})}{p(m_{x,y})}\frac{p(m_{x,y}|z_{1:j-1})}{p(z_j|z_{1:j-1})}
\end{equation}
where $p(m_{x,y}|z_j)$ is the inverse measurement model, $p(m_{x,y}|z_{1:j-1})$ is the previous measurement, and $p(m_{x,y})$ is the prior map.
In order to avoid difficult-to-calculate probabilities we use the binary Bayes filter in log odds form:
\begin{equation}
    l_{j}=l_{j-1} + \log\frac{p(m_{x,y}|z_j)}{1-p(m_{x,y}|z_j)} + \log\frac{p(m_{x,y})}{1-p(m_{x,y})}
\end{equation}
where $l_j = \log\frac{p(m_{x,y}|z_{1:j})}{1-p(m_{x,y}|z_{1:j})}$, the log odd form of the inverse measurement model ($\log\frac{p(m_{x,y}|z_j)}{1-p(m_{x,y}|z_j)}$) assigns to all cells within the 3D labelled points $\mathbf{X}_{j}$ an occupancy value $l_{occ}$. In the experiments $l_{occ} = 0.9$ has been taken into account.
Since we have no prior knowledge of the map $p(m_{x,y}) = 0.5$.
A cell is considered occupied by a rock if the probability exceed a threshold of 70\%. The output of the data fusion is a three dimensional semantic map where for each voxel only one label is associated. 


\section{Experimental Results}

\begin{figure}[ht]
	\centering
	\includegraphics[width=1\linewidth]{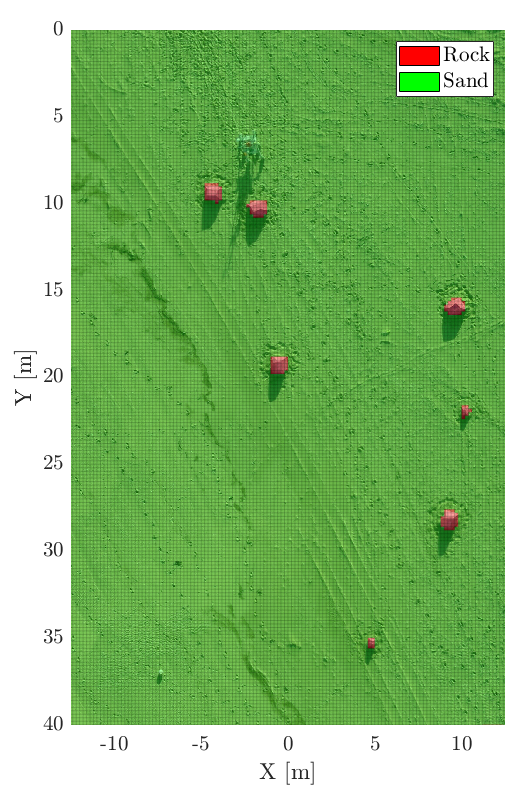}
	\caption{Ortomosaic taken by the drone with manually labelled rocks used as ground truth reference.}
	\label{fig::groundTruth_map}
\end{figure}

\begin{table}[htbp]
	\caption{Sensors used in this work (see \cite{hewitt2018katwijk}).}
	\begin{center}
		\begin{tabular}{|p{1.5cm}|p{3cm}|p{2.5cm}|}
			\hline
			
			\textbf{Sensor} & \textbf{Description}& \textbf{Data logged} \\
			\hline
			\hline
			LocCam & PointGery Bumblebee2 (BB2-08S2C-38) 12cm baseline stereo camera & 1032 $\times$ 776 images\\
			\hline
			RTK GPS & Trimble BD 970 Receiver with Zephyr Model 2 Antenna (rover) Trimble BX 982 Receiver with Zephyr Geodetic Antenna (base station)  & Latitude, Longitude and Altitude expressed on WGS 84 ellipsoid\\
			\hline
		\end{tabular}
		\label{tab::sensor_char}
	\end{center}
\end{table}

\begin{table*}[ht]
	\caption{Semantic segmentation 3D mapping performances in terms of Accuracy and IoU.}
	\begin{center}
		\begin{tabular}{|c|c|c|c|c|c|c|}
			\hline
			& \multicolumn{2}{|c|}{Data 01} & \multicolumn{2}{|c|}{Data 02} & \multicolumn{2}{|c|}{Data 03}\\\cline{2-7}
			& \textbf{Accuracy}& \textbf{IoU} &  \textbf{Accuracy}& \textbf{IoU} & \textbf{Accuracy}& \textbf{IoU}\\
			\hline
			\hline
			DeepLabv3+ & 0.96 & 0.58 & 0.90 & 0.21 & 0.95 & 0.36 \\
			\hline
			MLESAC & 0.95 & 0.37 & 0.97 & 0.58 & 0.96 & 0.47\\
			\hline
		\end{tabular}
		\label{tab::results}
	\end{center}
\end{table*}

The  metrological  performances  of the  proposed  method  have been evaluated using the  ESA  Katwijk  Beach Planetary Rover Dataset\footnote{https://robotics.estec.esa.int/datasets/katwijk-beach-11-2015/} \cite{hewitt2018katwijk}, which provides Mars analogous images. The dataset is composed by acquisitions from rover navigation sensors such as wheel odometry, IMU, Stereo LocCam, Stereo PanCam, ToF Cameras and LiDAR plus DGPS groundtruth data and the ortomosaic and DEM maps generated from aerial images taken by a drone. 
In this work, of the overall dataset, only the LocCam stereo images have been used for semantic maps generation. The characteristics of the sensors used in the experimental part are summarized in Table \ref{tab::sensor_char}. The DGPS has been used for registering the generated maps to the ortomosaic taken by the drone, which has been used as map referencing. Map ground truth has been obtained by manually labelling the ortomosaic, see \figurename{\ref{fig::groundTruth_map}}.

The  proposed  method  has been compared with a standard geometric point cloud processing algorithm for extracting objects out of a plane. This method is particularly effective on the tested dataset as the terrain is flat with the exception of protruding rocks. The algorithm uses MLESAC (Maximum Likelihood Estimation SAmple Consensus) \cite{torr2000mlesac}, for find the principal plane.

\begin{figure*}[htbp]
\centering
\subfloat[]{\includegraphics[height=3.5cm]{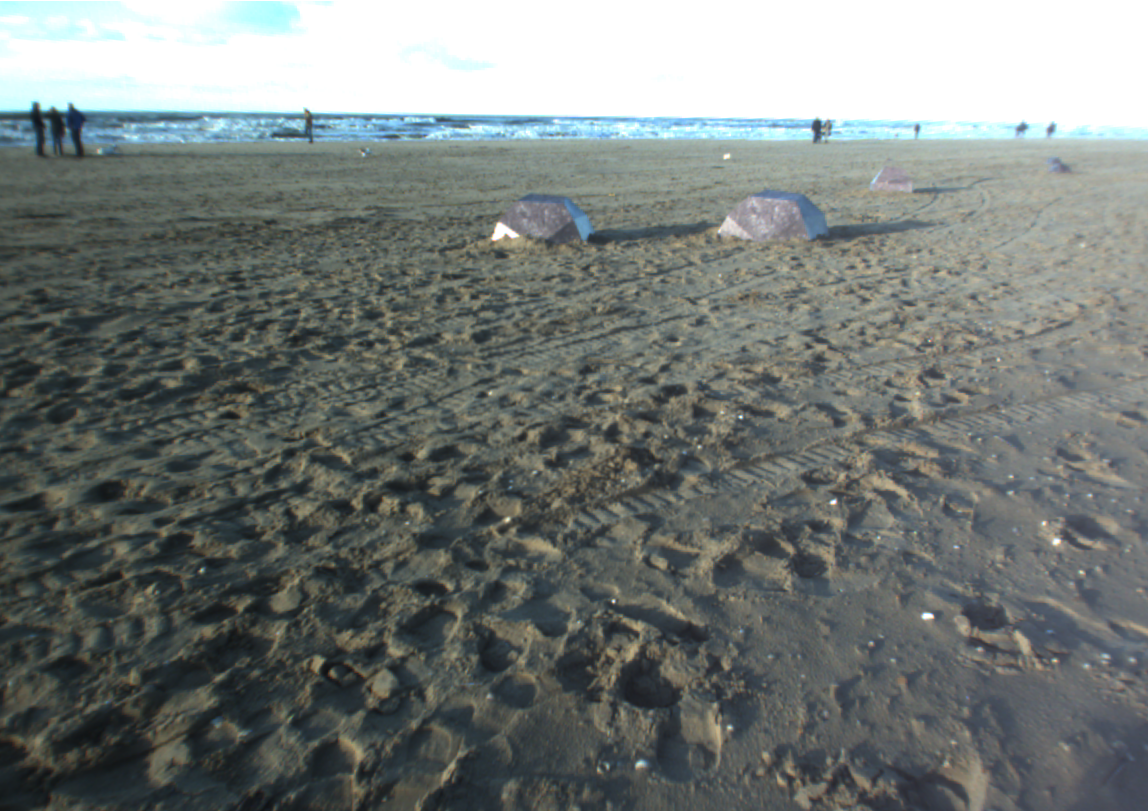}\label{fig::image}}
\hfill
\subfloat[]{\includegraphics[height=3.5cm]{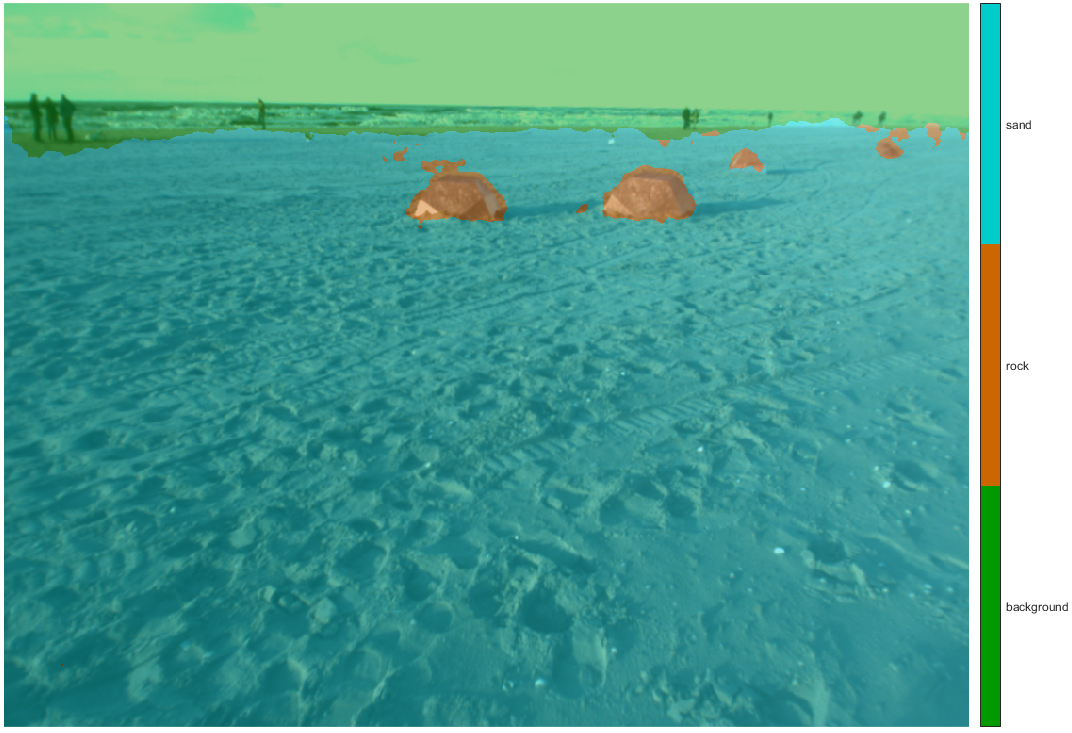}\label{fig::image_segmentation}}
\hfill
\subfloat[]{\includegraphics[height=3.5cm]{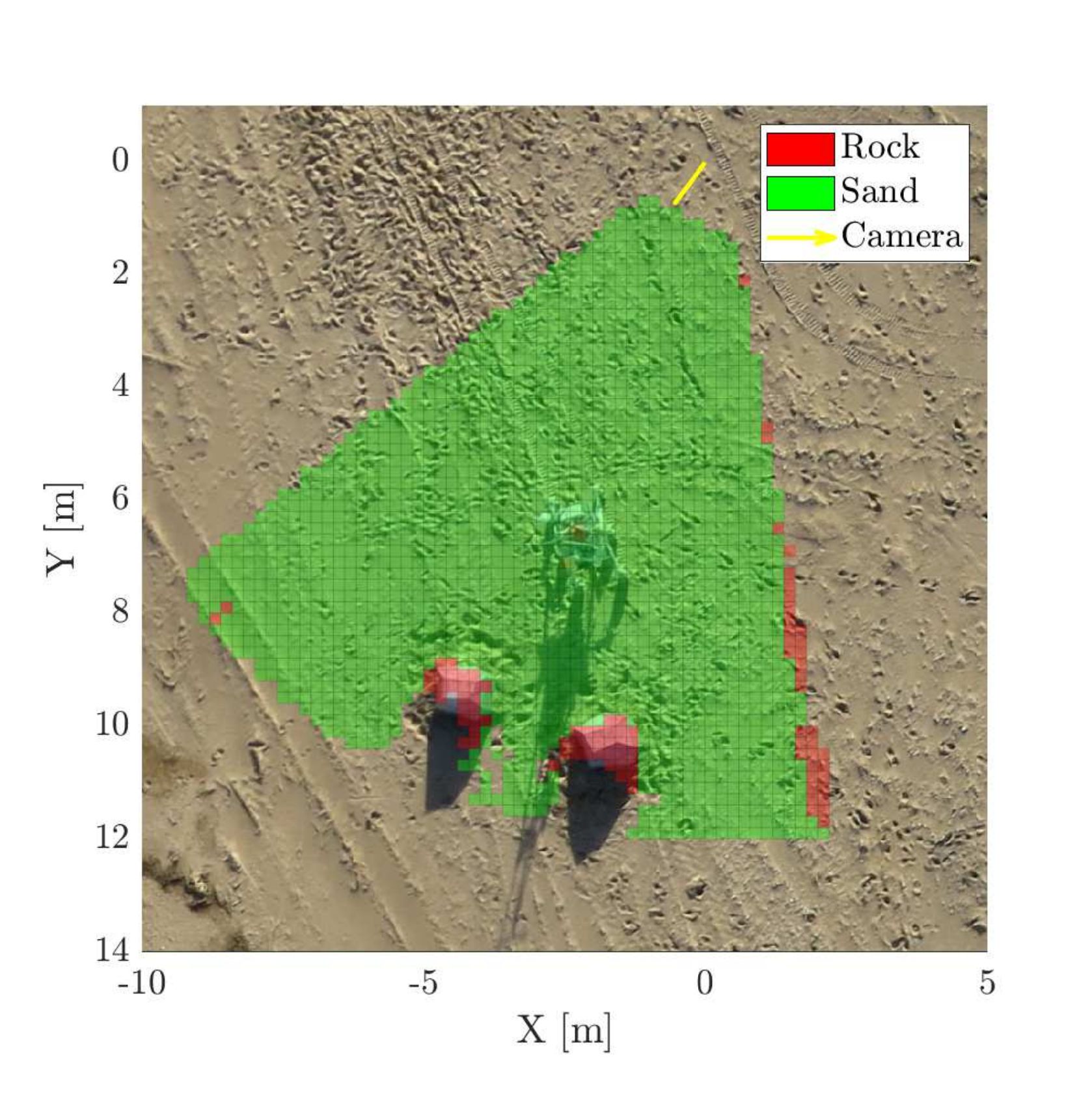}\label{fig::sem_seg_MSLEAC_01}}
\hfill
\subfloat[]{\includegraphics[height=3.5cm]{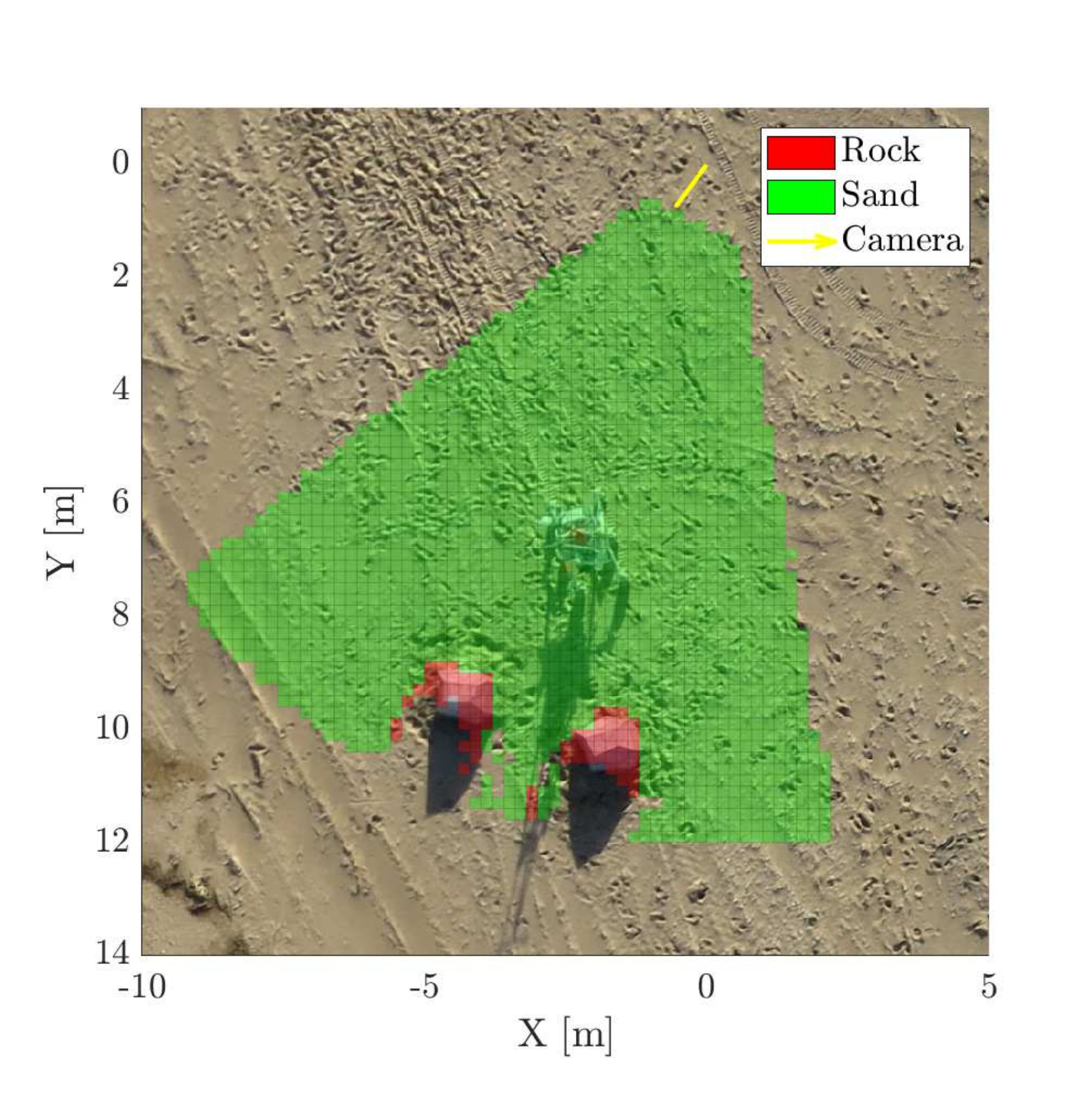}\label{fig::sem_seg_CNN_01}}
\hfill\\
\subfloat[]{\includegraphics[height=3.5cm]{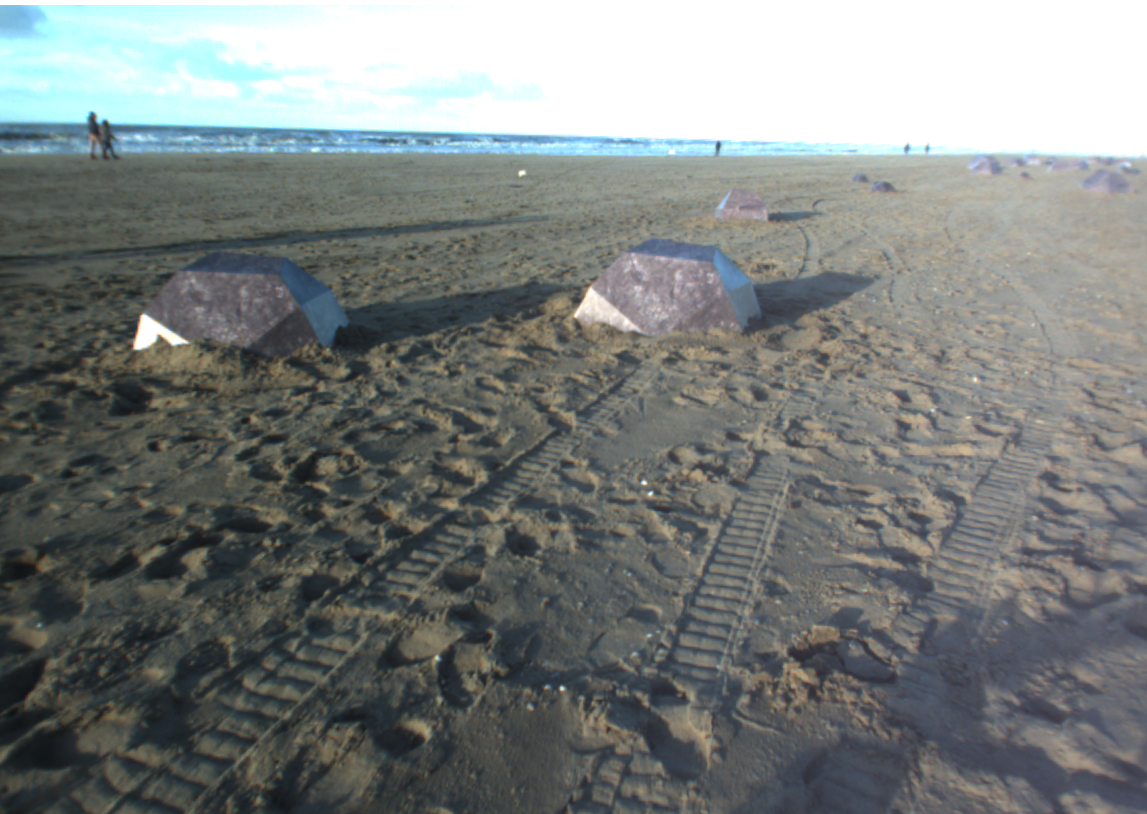}\label{fig::image_02}}
\hfill
\subfloat[]{\includegraphics[height=3.5cm]{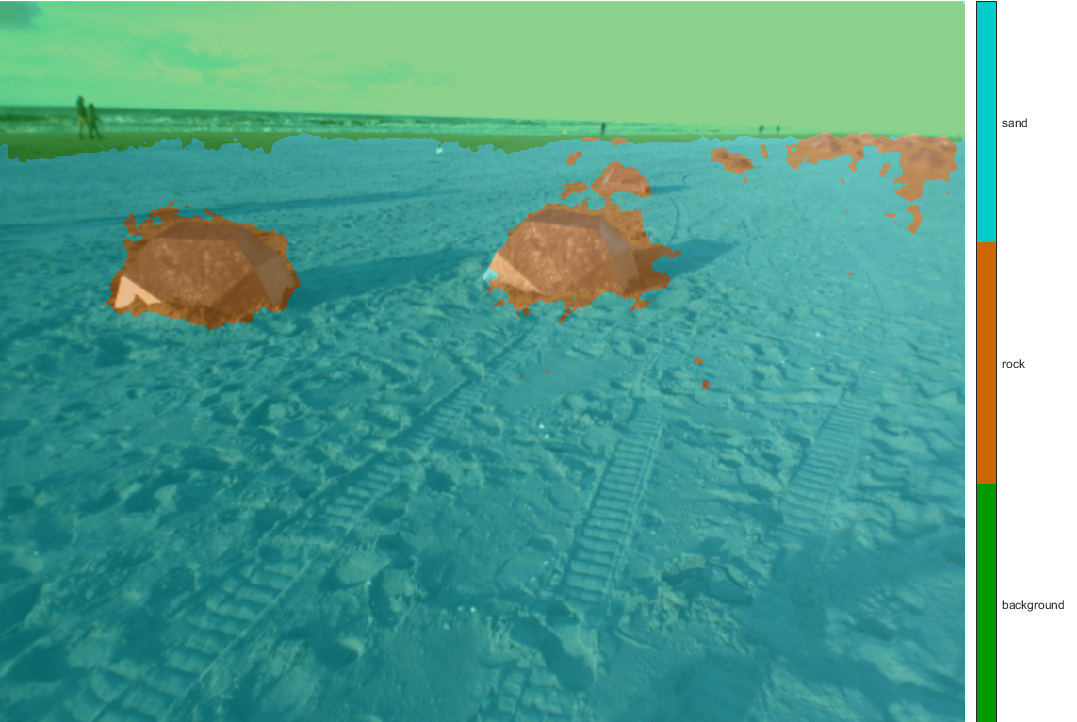}\label{fig::image_segmentation_02}}
\hfill
\subfloat[]{\includegraphics[height=3.5cm]{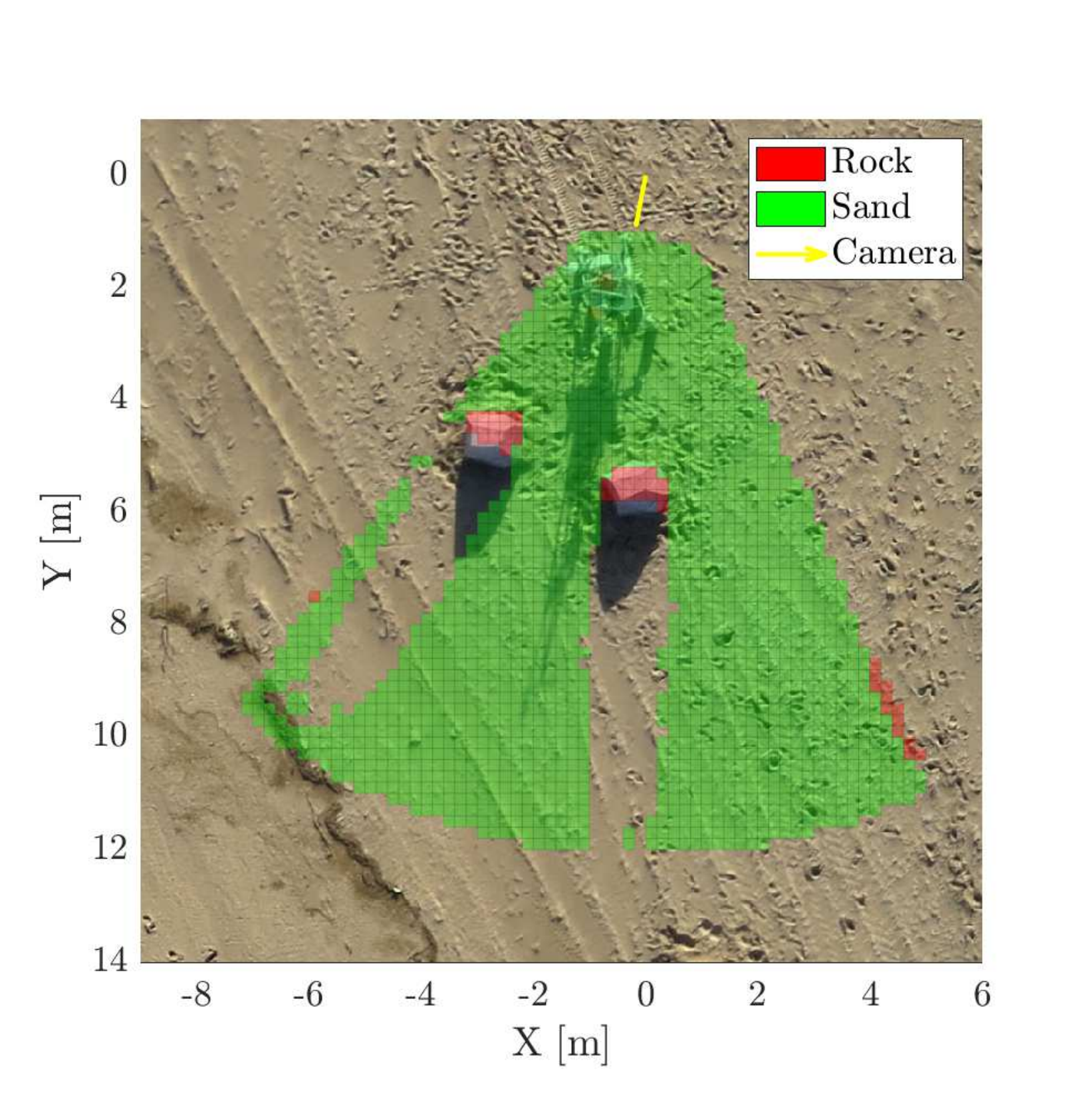}\label{fig::sem_seg_MSLEAC_02}}
\hfill
\subfloat[]{\includegraphics[height=3.5cm]{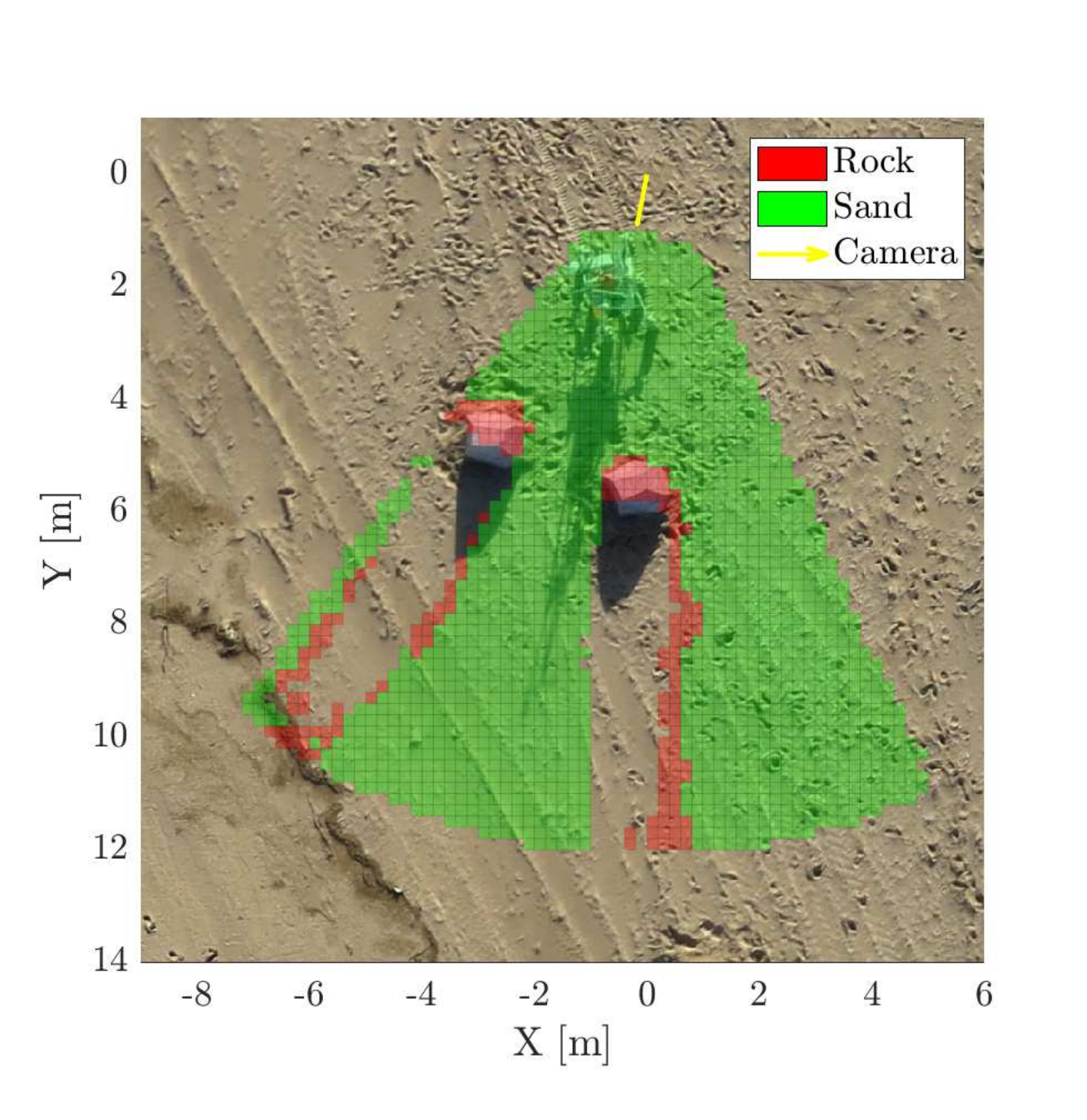}\label{fig::sem_seg_CNN_02}}
\hfill\\
\subfloat[]{\includegraphics[height=3.5cm]{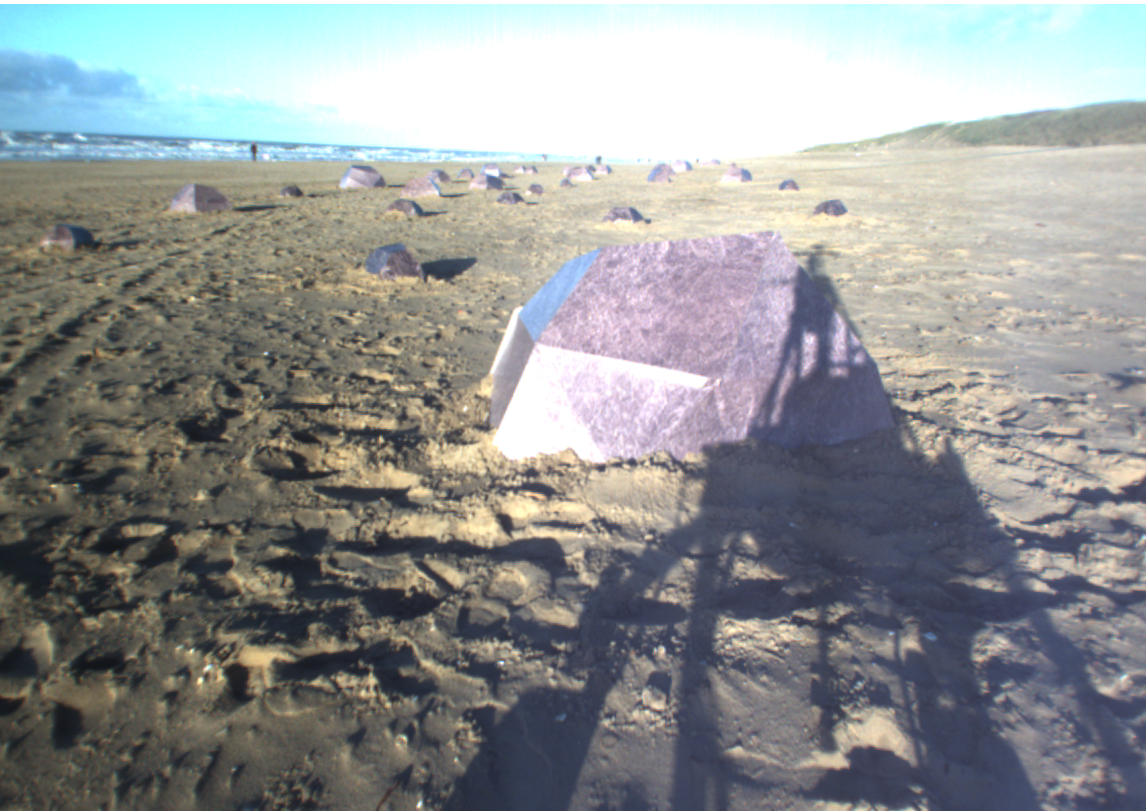}\label{fig::image_03}}
\hfill
\subfloat[]{\includegraphics[height=3.5cm]{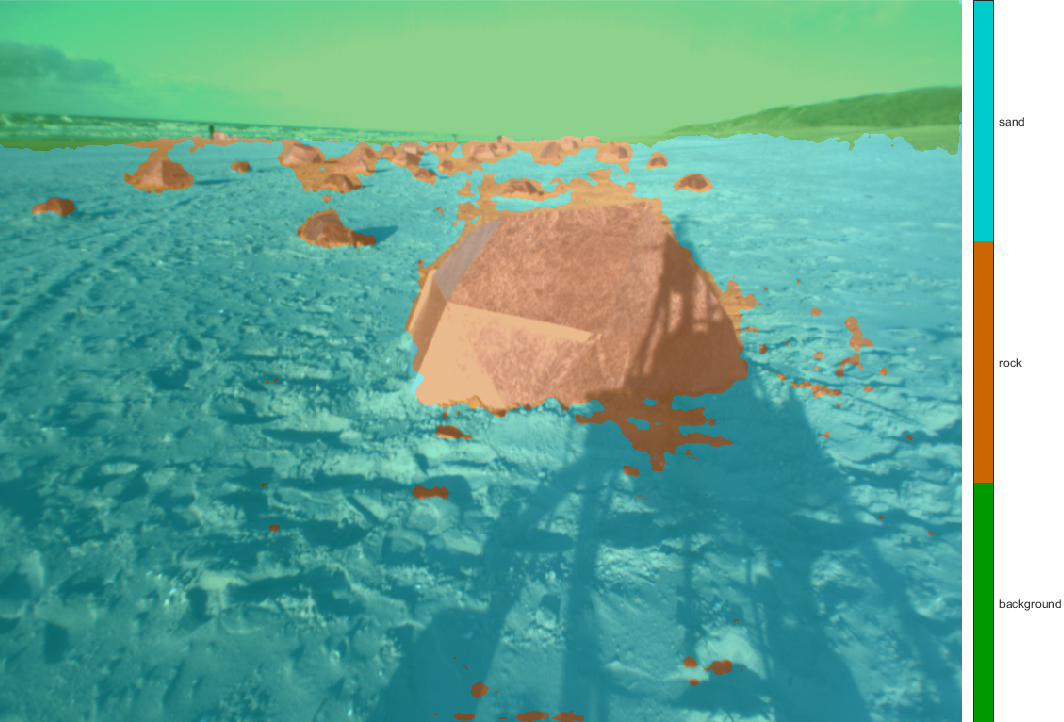}\label{fig::image_segmentation_03}}
\hfill
\subfloat[]{\includegraphics[height=3.5cm]{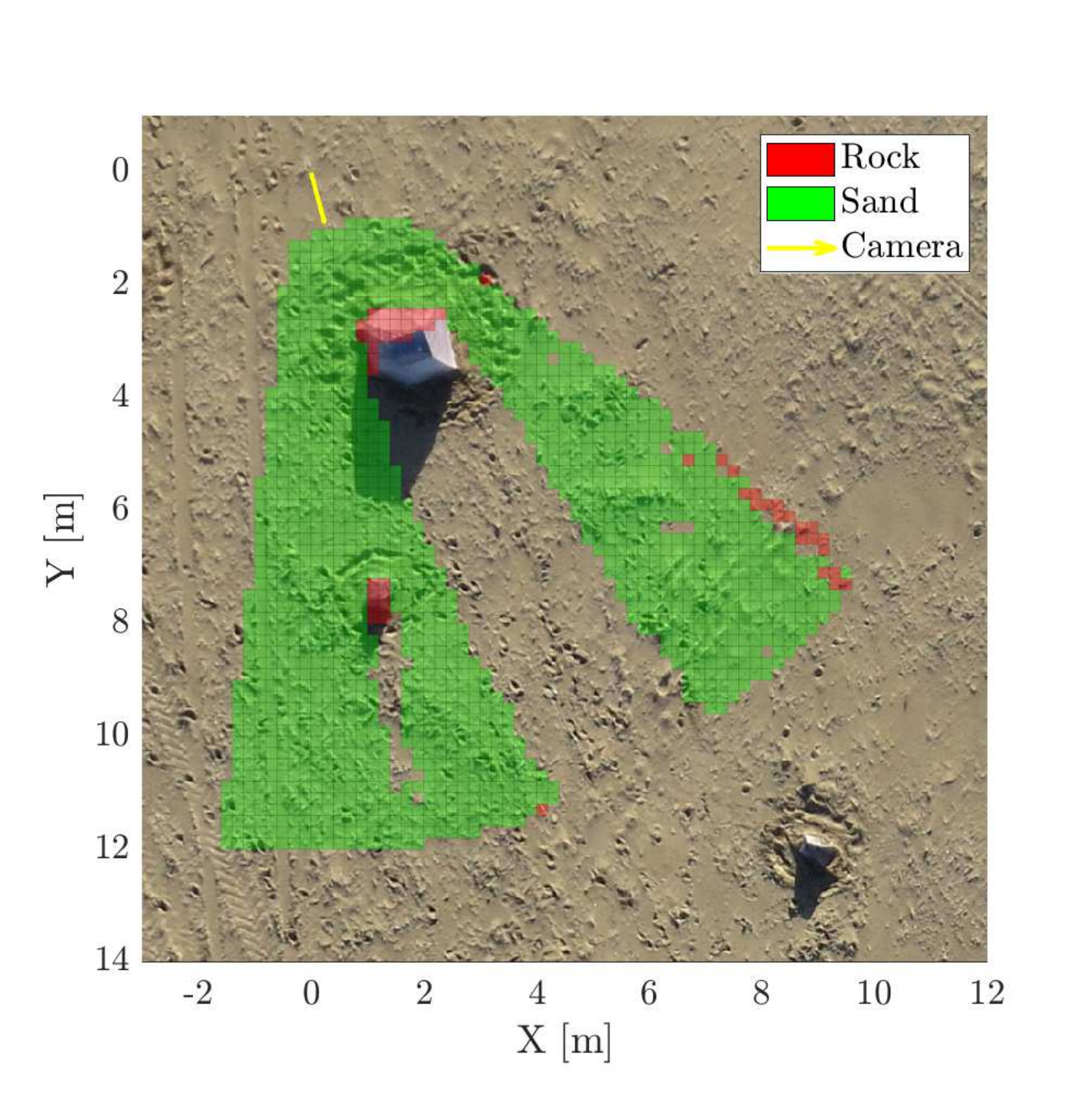}\label{fig::sem_seg_MSLEAC_03}}
\hfill
\subfloat[]{\includegraphics[height=3.5cm]{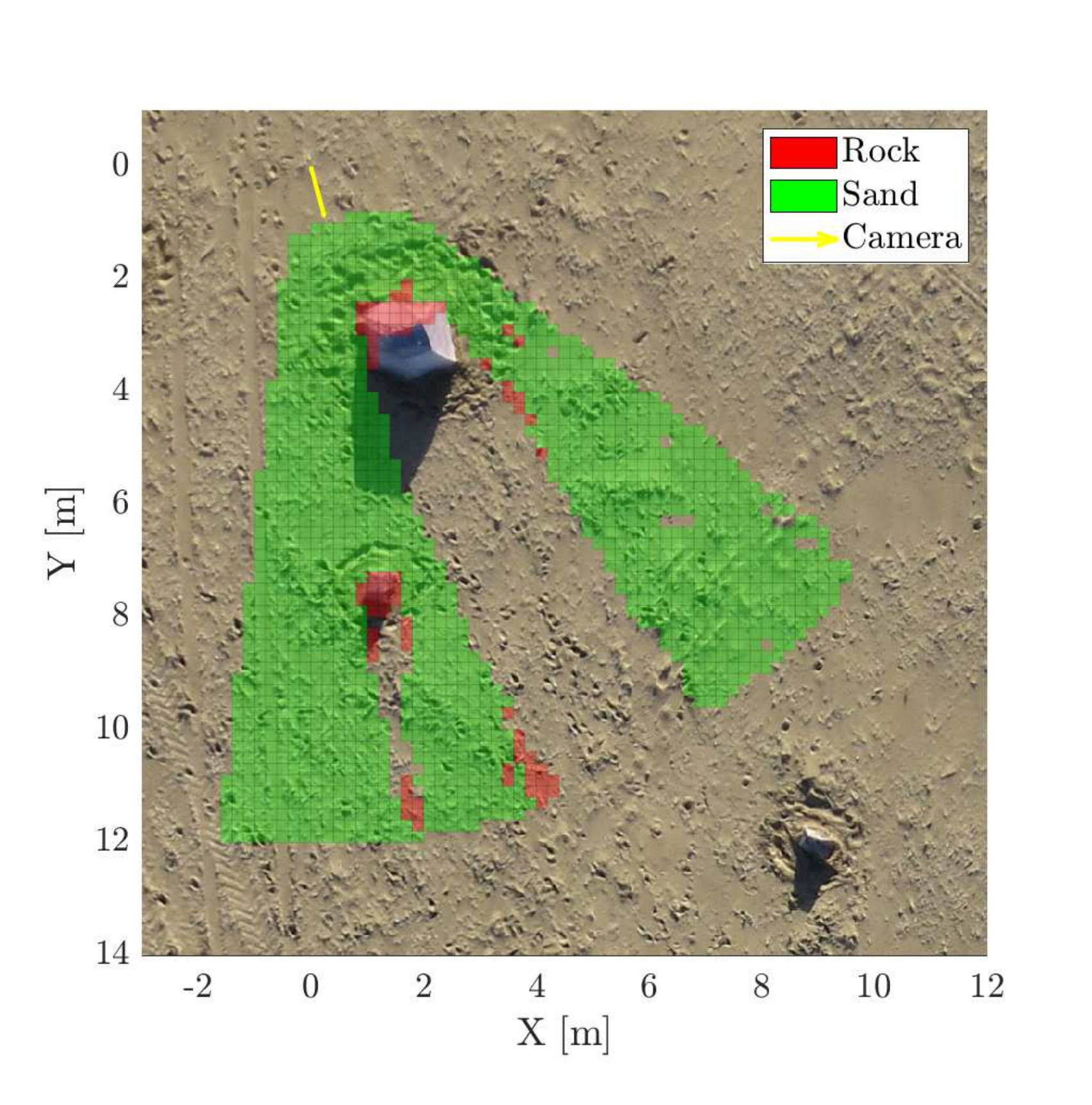}\label{fig::sem_seg_CNN_03}}
\hfill\\
\caption{Labelled images and generated semantic maps, first row: Data 01, second row: Data 02 and third row: Data 03. (a-c-i) Input image. (b-f-j) Labelled image using the   CNN  labeler  (DeepLabv3+). (c-g-k) Semantic map obtained with MLESAC plane segmentation. (d-h-l) Semantic map obtained with the proposed method.}

\end{figure*}

\figurename{\ref{fig::image}} (Data 01), \figurename{\ref{fig::image_02}} (Data 02) and \figurename{\ref{fig::image_03}} (Data 03) show some of the images for which we evaluated the segmentation algorithms performances. \figurename{\ref{fig::image_segmentation}}, \figurename{\ref{fig::image_segmentation_02}} and \figurename{\ref{fig::image_segmentation_03}} show images labelled with DeepLabv3+. As it is possible to observe in Data 01 the two rocky obstacles are few meter away from the camera and not easily recognizable, however they are correctly labelled by the CNN labeller. The labeller effectiveness can be observed also in \figurename{\ref{fig::image_segmentation_03}}: despite the fact that the rover's shadows is present both in the rock and in the sand it does not affect label performances. \figurename{\ref{fig::sem_seg_MSLEAC_01}}, \figurename{\ref{fig::sem_seg_MSLEAC_02}} and \figurename{\ref{fig::sem_seg_MSLEAC_03}} show the occupancy map obtained by filtering the MLESAC segmentation with the binary Bayes filter. \figurename{\ref{fig::sem_seg_CNN_01}}, \figurename{\ref{fig::sem_seg_CNN_02}} and \figurename{\ref{fig::sem_seg_CNN_03}} show the occupancy map obtained with the proposed method. It is possible to observe the presence of some false positive cells detected by the MSLEAC approach due to the point cloud noise. The CNN-based method is also subject to the presence of false positives, as shown in \figurename{\ref{fig::sem_seg_CNN_02}}, due to the tendency in some cases to label a slightly larger area.

In order to evaluate the method performances the semantic occupancy maps have been compared with a semantic map obtained from a manually labelled ortomosaic of the area taken by a drone. The following metrics have been used: accuracy and Intersection over Union (IoU):
\begin{equation}
    \mathrm{accuracy} =  \frac{\mathrm{TP} + \mathrm{TN}}{\mathrm{TP} + \mathrm{TN} + \mathrm{FP} + \mathrm{FN}}
\end{equation}
\begin{equation}
    \mathrm{IoU} =  \frac{\mathrm{TP}}{\mathrm{TP} + \mathrm{FP} + \mathrm{FN}}
\end{equation}
where TP represent the True Positive, TN the True Negative, FP the False Positive and FN the False Negative. Table \ref{tab::results} shows the proposed mathod semantic segmentation 3D mapping performances in terms of Accuracy and IoU. It is possible to see that the two methods have comparable performances in detecting the rocks, however the label extraction in the image is computationally more efficient than apply MLESAC for  plane  segmentation.

\section{Conlcusions}

In this paper we presented a terrain assessment method for Martian rover navigation based on Convolutional Neural Network labeling. The relatively new pre-trained network DeepLabv3+ has been used for semantic segmentation in the images space. The algorithm is capable to produce accurate three dimensional maps with associated label up to a dozen meters from the camera. The proposed method has been tested on public available dataset of a Martian analogous environment.   

\bibliographystyle{IEEEtran}
\bibliography{MAS2020_CNN_semanticMap}

\end{document}